\documentclass[9pt,sigconf]{acmart}

\AtBeginDocument{}

\copyrightyear{2023} 
\acmYear{2023} 
\setcopyright{acmlicensed}\acmConference[HotMobile '23]{The 24th International Workshop on Mobile Computing Systems and Applications}{February 22--23, 2023}{Newport Beach, CA, USA}
\acmBooktitle{The 24th International Workshop on Mobile Computing Systems and Applications (HotMobile '23), February 22--23, 2023, Newport Beach, CA, USA}
\acmPrice{15.00}
\acmDOI{10.1145/3572864.3580337}
\acmISBN{979-8-4007-0017-0/23/02}

\acmSubmissionID{71}

\usepackage{xspace}
\usepackage{enumitem}
\usepackage{tablefootnote}

\definecolor{pro_green}{rgb}{0.0, 0.66, 0.47}

\newcommand{\para }[1]{\medskip \noindent  {\bf #1}}

\newcommand{\frontfacing}{front-facing\xspace}
\newcommand{\backfacing}{back-facing\xspace}

\newcommand{\Frontfacing}{Front-Facing\xspace}
\newcommand{\Backfacing}{Back-Facing\xspace}

\setlist[itemize]{leftmargin=.12in}

\begin{document}

\title{Multi-Camera Lighting Estimation for Photorealistic Front-Facing Mobile Augmented Reality}

\author{Yiqin Zhao}
\orcid{0000-0003-1044-4732}
\affiliation{\institution{Worcester Polytechnic Institute}
 \streetaddress{100 Institute Road}
 \city{Worcester}
 \state{MA}
 \country{USA}
 \and
}
\email{yzhao11@wpi.edu}
\authornote{Work completed while part-time interning at Google.}

\author{Sean Fanello}
\orcid{0000-0001-9726-4501}
\affiliation{\institution{Google}
 \streetaddress{1600 Amphitheatre Parkway}
 \city{Mountain View}
 \state{CA}
 \country{USA}
 }
\email{seanfa@google.com}

\author{Tian Guo}
\orcid{0000-0003-0060-2266}
\affiliation{\institution{Worcester Polytechnic Institute}
 \streetaddress{100 Institute Road}
 \city{Worcester}
 \state{MA}
 \country{USA}
 }
\email{tian@wpi.edu}

\renewcommand{\shortauthors}{Zhao et al.}

\begin{abstract}
Lighting understanding plays an important role in virtual object composition, including mobile augmented reality (AR) applications. 
Prior work often targets recovering lighting from the physical environment to support photorealistic AR rendering. 
Because the common workflow is to use a \backfacing camera to capture the physical world for overlaying virtual objects, we refer to this usage pattern as \backfacing AR.
However, existing methods often fall short in supporting emerging \frontfacing mobile AR applications, e.g., virtual try-on where a user leverages a \frontfacing camera to explore the effect of various products (e.g., glasses or hats) of different styles.
This lack of support can be attributed to the unique challenges of obtaining $360^{\circ}$ HDR environment maps, an ideal format of lighting representation, from the \frontfacing camera and existing techniques. 
In this paper, we propose to leverage dual-camera streaming to generate a high-quality environment map by combining multi-view lighting reconstruction and parametric directional lighting estimation.
Our preliminary results show improved rendering quality using a dual-camera setup for \frontfacing AR compared to a commercial solution.
\end{abstract}



\maketitle

\section{Introduction}
\label{sec:introduction}

Lighting estimation is an important and long-standing task in computer vision and graphics communities~\cite{legendre-facelight,somanath-envmapnet,Garon2019}. Over the past few years, we have observed a wide adoption of lighting estimation methods for end-user-facing applications. 
For example, scene lighting estimation is essential to render visually-coherent virtual objects in mobile AR applications~\cite{somanath-envmapnet,zhao2022litar}; in computational photography, lighting estimation is useful because photos can be post-edited to have different lighting conditions~\cite{legendre-facelight}. 
In this work, we explore the key research questions in lighting understanding for an emerging application domain, \emph{\frontfacing mobile AR}, particularly try-on applications where end users leverage handheld mobile devices such as smartphones to overlay products of interest on their faces.
Mobile AR try-on apps promote online shopping and often require photorealism to provide user experiences on par with physical try-on.

\Frontfacing mobile AR try-on shares many practical challenges with other lighting estimation apps, e.g., limited field-of-views (FoVs), but also has its own unique challenges and opportunities.
For instance, even though commercial AR SDKs support the world-space device pose tracking via \backfacing cameras, they do not currently support such tracking via the \frontfacing camera. Adding such support is not a mere integration of \frontfacing camera streams to existing algorithms but requires solving challenges that are rooted in moving objects (face in this case) and limited overlaps between images.
On the other hand, \frontfacing mobile AR try-on is less sensitive to spatial variance because of the proximity of the observation (the phone camera) and the rendering (the face) positions~\cite{xihe_mobisys2021}. 
This insensitivity allows us to operate in the 2D image space---rather than the more computation-intensive 3D space~\cite{xihe_mobisys2021}---when generating the proper lighting information.

Providing a photorealistic and visually-coherent try-on experience can be boiled down to supporting reflection, shadow, and correct color tone~\cite{zhao2022litar}.
All three features can be achieved by utilizing a 360$^{\circ}$ HDR environment map~\cite{debevec2006image}.
However, obtaining such an environment map on mobile devices is difficult for key reasons, including small FoVs and limited support for HDR streaming.
Even in cases when high-end phones could capture HDR videos, they do not support \emph{simultaneous multi-camera streaming}. 
Furthermore, \frontfacing AR, compared to \backfacing AR, has an even more limited FoV, and the camera often fails to observe important environmental information that is useful for reflective rendering.
In short, it is challenging to obtain high-quality environment maps even on modern mobile devices.

This paper outlines a research roadmap that leverages \emph{multi-camera} to deliver high-quality lighting information for \frontfacing mobile AR.
State-of-the-art approaches either only support a subset of the features~\cite{zhao2022litar} or generate incoherent reflection~\cite{somanath-envmapnet}.
To overcome the limitations of prior work, our key idea is to generate a high-quality environment map by \emph{combining multi-view lighting reconstruction and parametric directional lighting estimation}.
Our system design uses LDR multi-camera video streaming as input because streaming dual-camera LDR images requires less energy, compared to HDR streaming, and is better supported across recent mobile devices.
Our proposed three directions are grounded on our preliminary investigation of dual-camera streaming (\S\ref{sec:the_promise_of_multi_view_cameras}). 
To efficiently increase mobile device observation, we propose to regulate the \backfacing camera usage by \emph{time and mobility} to conserve energy.
We propose a simple and lightweight face-tracking component for stitching camera views to generate an environment map. 
Lastly, we propose further improving the completeness and color tone of the environment map and supporting shadow casting by extending our parametric directional lighting estimation model with the \backfacing camera streaming. 

In summary, we make the following key contributions:

\begin{itemize}[leftmargin=.12in,topsep=4pt]
    \item We characterize the challenges of supporting photorealistic \frontfacing mobile AR, an important and emerging application domain of the lighting estimation task.
    \item We demonstrate that using a dual-camera setup can improve the mobile AR try-on experience with more visually-coherent rendering, especially for reflective objects.
\item We describe key questions and promising designs for achieving a visually-coherent virtual try-on experience in a dual-camera AR system, including reflection, accurate color tone, and shadow casting. 
\end{itemize} 
\section{Challenges}
\label{sec:motivation_for_photorealism}

\begin{figure}[t]
\centering
    \includegraphics[width=0.9\linewidth]{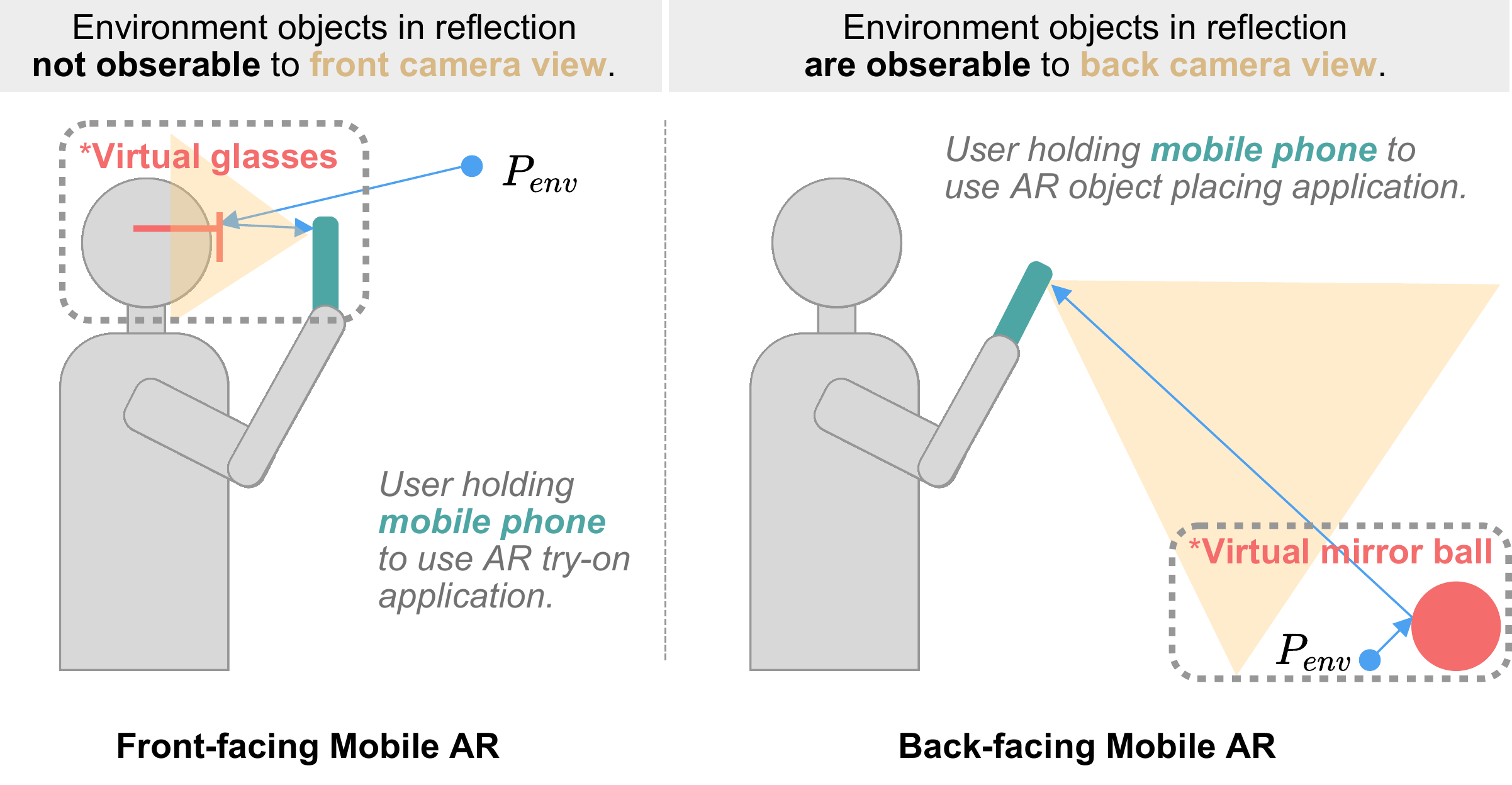}
    \caption{
    The challenge of reflective rendering.
\textnormal{
Mobile phones and example light rays are illustrated in green and blue, respectively.
}
    }
    \label{fig:reflection_directions}
    \vspace{-2mm}
\end{figure}

\Frontfacing mobile AR try-on, an emerging application of lighting estimation, exhibits new and unique challenges compared to other mobile AR shopping apps, e.g., IKEA Place\footnote{\url{https://tinyurl.com/yrruy65r}}. Because these AR apps often use the \backfacing camera, in this paper, we refer to them as \emph{\backfacing AR}.
Figure~\ref{fig:reflection_directions} illustrates the key usage difference between these two types of AR apps and the implication on reflective rendering. 
In the \backfacing AR, the user often points the mobile camera toward a rendering position; the environment information required for rendering is often included in the corresponding camera views.
In contrast, users often hold the mobile camera relatively closer to their faces when they use \frontfacing AR apps.
Because of the capturing direction of the \frontfacing camera, it often misses the relevant environment information, making reflection rendering more challenging.

\begin{table}[t]
   \caption{
        Comparisons of AR framework lighting estimation support for \frontfacing mobile AR apps.
        \textnormal{
}
    }
  \vspace{-2mm}

  \label{tab:ar_framework_comparision}
  \scriptsize \centering \begin{tabular}{lrrrr}
  \toprule
   \textbf{Framework} &  
   \textbf{Ambient} &
   \textbf{Directional} &
   \textbf{Spherical Harmonics} &
   \textbf{Env. Map} \\
\midrule
    ARKit\tablefootnote{\url{https://developer.apple.com/augmented-reality/}} & Yes & Yes &  Yes & No \\
    ARCore\tablefootnote{\url{https://developers.google.com/ar}} & Yes & No &  No & No \\
    Proposed & Yes & Yes &  Yes & Yes \\
  \bottomrule
  \vspace{-5mm}
  \end{tabular}\end{table}

Existing mobile AR-based try-on apps leverage depth estimation, 3D face modeling~\cite{paysan-baselfacemodel}, and real-time rendering to improve object fitting and pose tracking.
However, due to the usage of reflective materials in many product designs, visual incoherence can occur if the 3D product objects are not rendered with the correct environmental lighting.
At a high level, environment lighting requires a complex data structure representing omnidirectional environment observations and directional intensity variations.
Traditionally, high-quality environment lighting can be obtained using specialized hardware like physical light probes or 360$^\circ$ cameras, which are cumbersome to be integrated into mobile AR apps.
Table~\ref{tab:ar_framework_comparision} shows that existing mobile AR frameworks only provide partial lighting support and cannot effectively leverage the new features on modern mobile phones like multi-camera streaming.
In contrast, our proposed framework will provide a comprehensive set of lighting-related features for \frontfacing mobile AR.

\section{The Promise of Multi-Camera}
\label{sec:the_promise_of_multi_view_cameras}

We show that dual-camera streaming, a recent  hardware advancement supported by major phone vendors, can improve the visual results for \frontfacing AR-based try-on.
Furthermore, we show that EnvMapNet~\cite{somanath-envmapnet}, a state-of-the-art lighting estimation model, cannot fully utilize the dual-camera inputs.
Additional setup details are described below, and the research challenges and proposed solutions are presented in \S\ref{sec:research_roadmap}.

\begin{figure*}[t]
\centering
    \includegraphics[width=\linewidth]{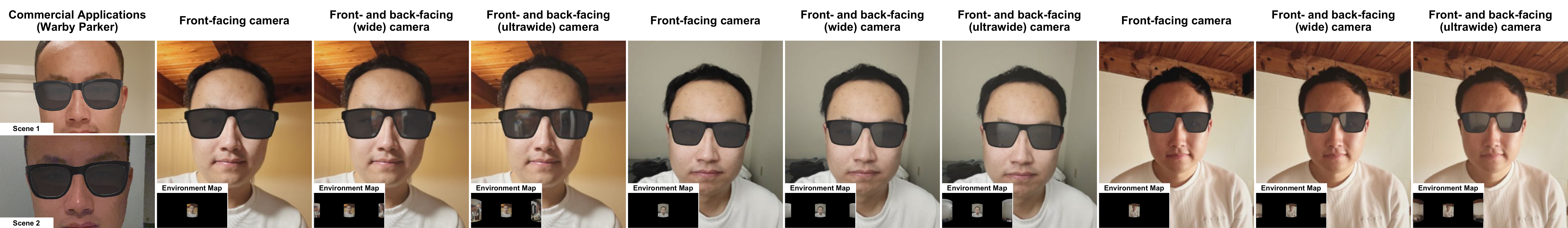}
    \caption{Reflection rendering comparisons of different rendering setups.
        \textnormal{
        Warby Parker uses a static environment map which leads to incorrect reflections. 
        Switching from \frontfacing to dual-camera streaming results in a more complete environment map and improved reflection rendering.
}
    }
    \label{fig:visuals}
    \vspace{-3mm}
\end{figure*}

\para{Setup.}
We used a tripod to fix the mobile phone position and then captured the \frontfacing and \backfacing camera images, for three indoor scenes. 
Specifically, we used the \frontfacing, wide, and ultra-wide \backfacing cameras, with 70$^{\circ}$, 75$^{\circ}$, and 120$^{\circ}$ FoVs, on iPhone 14 Pro.
Next, we imported the captured images into \texttt{Blender} and generated virtual sunglasses on top of the user's face using the \frontfacing camera image.
We tested the energy usage of multi-camera streaming using Apple's sample dual-camera capturing code\footnote{\url{https://tinyurl.com/mr32e7ab}}.
We measured the device battery usage by running the dual-camera streaming app for 10-minute sessions on an iPhone 11 Pro.
We set the screen brightness value to 50\% and disabled all background tasks and networking, and measured the idle device battery consumption. 
After each measurement, we recharged the battery to 90\%.
We tested 6 configurations by keeping the \frontfacing camera always on and turning on/off the \backfacing camera programmatically.
We repeated the measurement 3 times for each configuration and reported the average value.

\begin{figure*}[t]
\centering
    \includegraphics[width=0.9\linewidth]{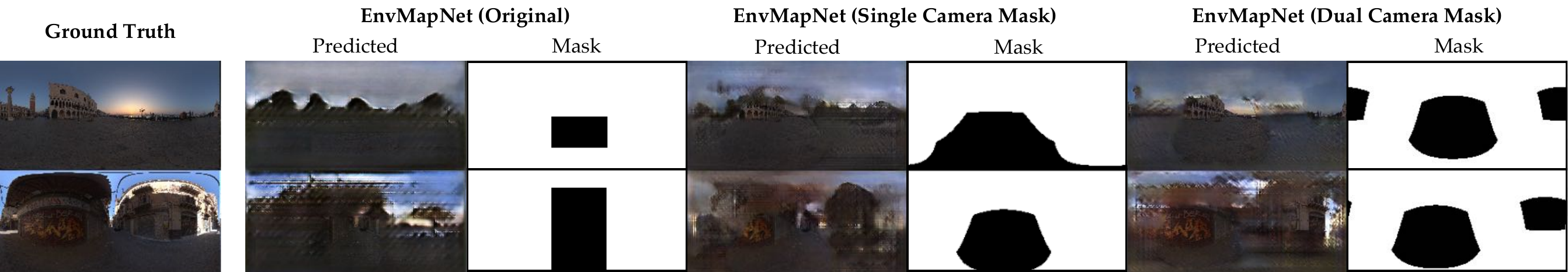}
    \caption{
        Visual comparisons of EnvMapNet~\cite{somanath-envmapnet} with different masks.
        \textnormal{We apply different masks to simulate different camera view setups. 
        In addition to EnvMapNet~\cite{somanath-envmapnet}'s original mask inputs, we provide two mask setups with 70$^\circ$ front-camera FoV and a 120$^\circ$ back-camera FoV. 
        We observe that even though dual-camera improves the prediction accuracy, the environment map details and light source directionalities are still inaccurate.}
\textnormal{
}
    }
    \label{fig:envmapnet}
\end{figure*}

\subsection{Rendering Result Analysis}
\label{subsec:rendering_visual_impacts}

Figure~\ref{fig:visuals} shows the rendering visual results using different camera setups and a commercial try-on app  Warby Parker~\footnote{\url{https://www.warbyparker.com}}.
We observe that combining \frontfacing and \backfacing camera views improves the environment map's quality and, consequently, the rendering quality.
Specifically, the rendered sunglasses in the camera setup with the largest FoV, i.e., \emph{\frontfacing and (ultra-wide) \backfacing cameras}, are more visually realistic and coherent than the other two renderings.

Capturing high-quality ground truth lighting information is often challenging and requires using specialized hardware~\cite{reinhard:2010:high}.
Prior work used a sphere metal ball to recover lighting information. Still, the resulting environment map can often show distortion and lower quality (compared to direct camera capture)~\cite{zhao2022litar}. 
In this work, we use the rendering results from the \emph{\frontfacing and (ultra-wide) \backfacing cameras} as the basis to calculate the PSNR. To avoid the impact of non-virtual objects on the calculation, we only calculate the PSNR of rendered virtual sunglasses between different camera setups. 
For front-only and front+wide \backfacing camera setups, we calculated the average PSNR of 25.70db and 27.05db, respectively.
The results align with the visual appearances and indicate the lower qualities of the \frontfacing camera setup. 

To quantify the environment map's completeness, we used a metric from our prior work that calculates the environment observation coverage over uniformly sampled directions~\cite{xihe_mobisys2021}.
Assume that the \frontfacing and \backfacing wide cameras share the same origin and the image ratio of 4:3. 
This setup covers 22.81\% of the observation directions, 2.14x of using the \frontfacing camera only.Switching from the wide to the ultra-wide camera leads to 38.28\% direction coverage, 3.6x of using \frontfacing camera alone.

\subsection{Evaluating the Performance of EnvMapNet}

We evaluated EnvMapNet's ability to generate HDR environment maps for dual-camera setup (dual camera mask) by training the model following the hyper-parameters and training setting described in the paper~\cite{somanath-envmapnet}. 
The masks are generated assuming a 70$^\circ$ front-camera FoV and a 120$^\circ$ back-camera FoV. 
We constructed a dataset using 450 online free HDR environment maps and 2100 data items from the Laval HDR dataset~\cite{Gardner2017}.
We observe that the dual-camera setup improves the average PSNR of predicted environment maps from 10.3db/11.7db (original/single camera mask) to 12.1db. 
Figure~\ref{fig:envmapnet} shows an example of visual comparisons.  
As we can see, even though dual-camera improves the prediction accuracy, the environment map details and light source directionalities are still inaccurate.

\begin{table}[tb]
   \caption{
        Battery usage for dual-camera streaming.
        \textnormal{
}
    }
  \vspace{-2mm}
    
  \label{tab:battery_measurement}
  \scriptsize \centering \begin{tabular}{llr}
  \toprule
   \textbf{\Frontfacing (mins)} &  
   \textbf{\Backfacing (mins)}  &
   \textbf{Battery Usage} \\
  \midrule
    0 & 0 & <1\% \\
    10 & 0 & 2\% \\
    10 & 1 & 2\% \\
    10 & 3 & 3\% \\
    10 & 5 & 4\% \\
    10 & 7 & 5\% \\
    10 & 10 & 5\% \\
  \bottomrule
  \end{tabular}\vspace{-5mm}
\end{table}

\subsection{Energy Consumption Measurement}
Table~\ref{tab:battery_measurement} shows the relationship between various dual-camera streaming configurations and battery usage. 
For a 10-minute session, simultaneously streaming \frontfacing and \backfacing cameras takes on average 5\% of total battery, 2.5x of capturing video using the \frontfacing camera alone.
When reducing the \backfacing camera capturing time to 3 minutes, the total battery usage drops to 3\%.
This suggests that selectively turning on and off the \backfacing camera when it is not needed can conserve battery. 
This observation provides an important insight for our adaptive dual-camera capturing policy design, as will be described in \S\ref{subsec:simultaneous_front_and_back_view_capturing}.
Further, our testing app consumes 62.7MB of memory in the \frontfacing-only mode while using 78MB of memory when streaming with both cameras, a 24.4\% increase.
Our empirical observation aligns with a recent work that leverages two \backfacing cameras to remove face motion blur~\cite{lai-facedeblurring}.

\begin{figure*}[t]
\centering
    \includegraphics[width=0.9\linewidth]{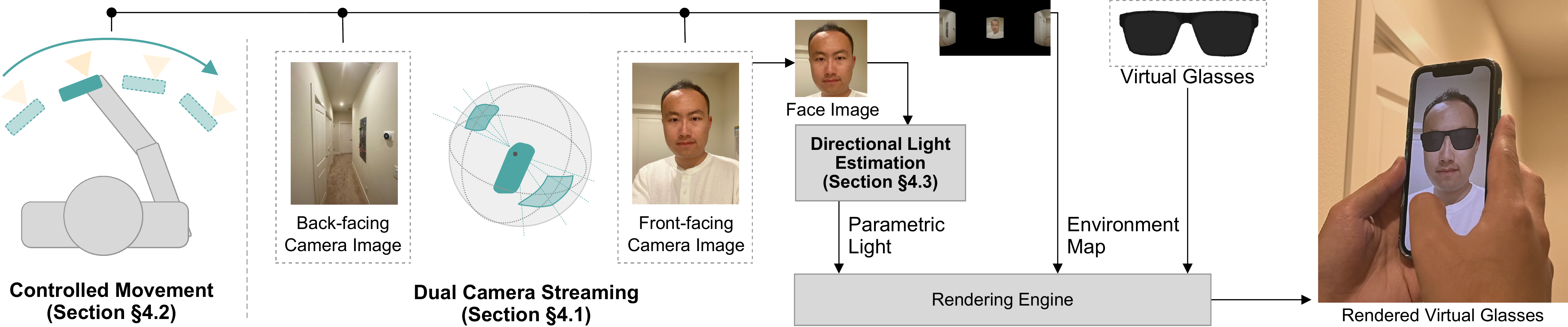}
    \caption{Example workflow.
        \textnormal{
            Our proposed system fuses dual-camera images to generate high-quality environment maps (\S\ref{subsec:simultaneous_front_and_back_view_capturing}) by leveraging a controlled movement pattern (\S\ref{subsec:multi_frame_environment_map_reconstruction}) to stitch multi-view images and  a parametric directional lighting model to improve the environment map's completeness and support correct color tone and shadow (\S\ref{subsec:parametric_directional_light_estimation}).
}
}
    \label{fig:applicaiton_workflow}
    \vspace{-3mm}
\end{figure*}

\section{Research Roadmap}
\label{sec:research_roadmap}

This section presents three key questions and promising designs for achieving a visually-coherent virtual try-on experience in \emph{a dual-camera AR system}, including reflection, accurate color tone, and shadow casting.  
Figure~\ref{fig:applicaiton_workflow} illustrates how solutions to these questions could improve the existing \frontfacing mobile AR workflow.
\begin{itemize}
    \item \textbf{Q1:} How to efficiently increase the mobile device observation?
The ideal solution should fully utilize the existing mobile hardware in an energy-conserving way.  
    \item \textbf{Q2:} How to effectively fuse camera observations to generate an environment map? The ideal solutions should handle temporal visual data in a noisy environment. 
    \item \textbf{Q3:} How to leverage learning-based models to estimate out-of-camera views and support shadow casting? The ideal solution should work nicely with dual-camera LDR image inputs.
\end{itemize}

\subsection{Dual-Camera Streaming Policies}
\label{subsec:simultaneous_front_and_back_view_capturing}

Based on our preliminary investigation that dual-camera streaming can increase the observation, but at the expense of mobile energy (\S\ref{sec:the_promise_of_multi_view_cameras}), we propose to regulate the \backfacing camera usage by \emph{time and mobility} to conserve energy.
Our proposed two policies are extended from our recent work LitAR's noise-tolerant data capturing policies for \backfacing AR~\cite{zhao2022litar} and are inspired by the adaptive policy from Lai et al. for deblurring~\cite{lai-facedeblurring}.

First, we propose to use a simple timer-based policy to turn on and off the \backfacing camera to capture potential environmental changes. 
The timer starts at a larger interval $w$, e.g., more than the rendering interval of 33 ms, and will be adjusted based on observed changes. 
Once the \backfacing camera is on, we will stream for $w$ before turning it off to mitigate the tail energy phenomenon~\cite{Huang2012-yq}.
We will decrease the timer interval by half if we observe sufficient environmental changes, indicating a more dynamic scene, over the capturing window. 
We could effectively determine the change between two consecutive images with metrics such as SSIM or PSNR. 
If little change is observed, we will gradually increase the timer interval. 
This policy is similar to the additive increase/multiplicative decrease algorithm.

Second, we propose to leverage the inertial measurement unit (IMU) sensor data to trigger \emph{early capturing} when significant device movement is detected.
To avoid capturing images with an undesirable visual artifact, e.g., motion blur, the \backfacing camera will only start streaming once the mobile device is stable.  
Every motion-triggered capturing will reset the timer.

\subsection{LDR Environment Map Generation}
\label{subsec:multi_frame_environment_map_reconstruction}

Obtaining 360$^{\circ}$ HDR images for visually coherent lighting rendering is a challenging task for mobile AR.
Even with the dual-camera streaming strategies described above, obtaining the 360$^{\circ}$ HDR environment map is still challenging because modern mobile cameras only have limited pixel intensity range support.
Consequently, we propose to use LDR environment maps and parametric directional lighting model (\S\ref{subsec:parametric_directional_light_estimation}) to approximate the HDR environment maps.
This section describes our proposal to leverage guided user movement to combine multi-view images (both \frontfacing and \backfacing) to generate LDR environment maps.

\para{Image-Based Stitching Via Face Tracking.}
We propose a simple and lightweight component to exploit device mobility and the dual-camera streaming feature to obtain high-quality device pose tracking data and generate high-quality environment maps.
The key idea is to use a controlled movement pattern and leverage the user's faces in \frontfacing camera images to establish \emph{a shared anchor point}, which can then be used as the reference for tracking device poses across frames. 

Our design is driven by the advancement of 3D face modeling research.
Using face geometry and camera parameters estimation models, e.g. MediaPipe\footnote{\url{https://mediapipe.dev}} face mesh~\cite{grishchenko2020attention}, we could obtain the relative pose of the camera to the user's face.
To combine multi-frame camera images, we propose to use the user's face as a shared anchor point by asking the user to fix their head pose during capturing.
To ease the user experiences, we propose to design an onscreen instruction for guiding the users to keep their head static while moving the mobile device along the trajectory of the great circle (with the radius of the user's arm length)\footnote{Guided movement is a common design for mobile applications such as taking panorama~\cite{kan2015interactive}.}. 
Figure~\ref{fig:applicaiton_workflow} illustrates the idea of controlled movement.
We only ask the user to perform this guided movement at the app launch time and when significant movement is detected while using the app. 

\para{Stitching Image Views for Environment Maps.}
To leverage the additional environment observations for improving reflection rendering quality, we first stitch the front and back camera images into a partial 360$^\circ$ environment map. 
Specifically, we assume the \frontfacing and \backfacing cameras share the same origin and map the camera image pixels to an environment map based on camera physical parameters and the pinhole camera model~\cite{wiki-pinholecamera}.
To bootstrap the user experiences when the AR application starts, we initialize the environment map data from the guided control movement at the beginning of the try-on app.
During application usage, we progressively update the environment map using image feature matching methods, e.g., \cite{harris1988combined}, and merge new \backfacing camera images into the reconstructed environment map.
For \frontfacing camera image, as the user's head pose and facial expressions might change during the app usage, we propose to only use the current frame for environment map reconstruction.

\para{Analysis of Environment Observation.}
We present an analysis of the environment observation increase when combing multi-frame images.
To test the observation increase, we manually created a setup in Blender with 3 \backfacing camera images, each with 120$^{\circ}$ FoV.
We set the image pose along a horizontal movement trajectory to simulate a $\pm$30$^{\circ}$ movement.
We used the same observation coverage measurement metric as in \S\ref{sec:the_promise_of_multi_view_cameras} and found that the environment observation can be increased to cover 50\% directions, 1.3x of using one \backfacing camera image alone.

\subsection{Parametric Directional Light Estimation}
\label{subsec:parametric_directional_light_estimation}

HDR environment maps are commonly used for rendering shadows of virtual objects.
However, directly reconstructing 360$^{\circ}$ HDR environment maps requires prolonged user engagement for scanning and can be prone to quality issues.
Furthermore, our prior work shows that using learning-based models to inpaint camera views to 360$^{\circ}$ HDR environment maps can be highly inaccurate for complex real-world indoor scenes~\cite{zhao2022litar}.
To address these limitations, we propose to approximate the HDR environment maps by combining reconstructed LDR environment maps and estimated parametric directional lights.
At a high level, our proposed directional light estimation model will work in tandem with the environment map reconstruction pipeline described in previous sections. 
The estimation model will improve the prediction of the color tone and the completeness of the environment map, as well as supporting shadow rendering.

Specifically, we propose to extend the estimation model~\cite{legendre-facelight} to use the \backfacing camera images to improve the lighting color estimation accuracy.
We expect that this design will improve the estimation accuracy of the environment color even when unseen objects or unnatural environmental lighting appear in the image.
This is because \backfacing camera images greatly enriches the environment lighting information in the input data, leading to more accurate light color estimation.
Besides parametric directional lights, our estimation model also outputs a low-resolution environment map.
We plan to use the estimated environment map to fill unseen regions on the reconstructed environment map.
This design will further improve the reflection rendering and overall color tone accuracy.
As the reconstructed environment map is refined progressively during AR app usage, we expect the completeness of the environment map and directional light color to also be improved.

\section{Related Work}
\label{sec:realted_work}

\para{Lighting Estimation.} 
Recent deep learning-based work has investigated recovering lighting information in both outdoor and indoor scenes~\cite{Garon2019}, leverages images of arbitrary scenes, portraits~\cite{legendre-facelight}, and shadows~\cite{liu2020arshadowgan}.
Although DL-based approaches can often deliver good visual appearances, they require training on large datasets with the lighting ground truth and can have a longer than hundreds of milliseconds inference time~\cite{zhao2020pointar}.
Prior mobile-oriented work, including our own~\cite{xihe_mobisys2021,zhao2022litar}, addressed the above two issues by either forgoing the DL route~\cite{prakash2019gleam,zhao2022litar} or exploring the advancement in the mobile camera system for the \backfacing mobile AR applications~\cite{cheng2018learning,xihe_mobisys2021,somanath-envmapnet}. 
Our proposed work combines a promising set of existing lighting estimation techniques to deliver tailored experiences for mobile AR try-on applications.

\para{Virtual Try-On Applications.}
Recent work on virtual try-on has demonstrated good progress in overlaying clothes on a human body with little deformation and correct occlusion~\cite{meng2010interactive}.
Earlier work on image-based try-on takes two inputs, images of a person and the desired clothes, and generates a new image where the same person is now wearing the desired clothes~\cite{han2018viton}. Other fashion items, such as shoes, cosmetics, and glasses, have recently received more attention~\cite{kim2008adoption}. 
For example, ARShoe proposed a real-time mobile AR-based shoe try-on system~\cite{an2021arshoe} that improved upon PIVTONS~\cite{chou2018pivtons}.
Despite the growing interest in virtual try-on applications, very few works address the visual coherency of the rendered fashion items, which is the focus of our work.

\para{Multi-Camera Mobile System.}
To address the limitation of low FoV associated with a single camera, several works have proposed to fuse the inputs from multiple cameras to improve the quality of lighting estimation. 
GLEAM supported an optional feature that allows different mobile devices (hence cameras) to share lighting information~\cite{prakash2019gleam}.
Li et al. proposed to use a $360^{\circ}$ panoramic stereo installed in the targeted indoor scene to estimate the scene properties, including lighting~\cite{li2021lighting}.
Multi-camera systems have also been proposed for other tasks, such as removing face blurs~\cite{lai-facedeblurring} and estimating depth~\cite{zhang20202}. 
Our proposed work shares the key idea of leveraging multiple cameras to increase the scene observations but tackles three unsolved challenges for \frontfacing mobile AR.

\section{Conclusion}
\label{sec:conclusion}

Lighting estimation has attracted a lot of attention over the past few years to improve the realism and visual details of AR apps~\cite{zhao2020pointar,Garon2019,Gardner2017,xihe_mobisys2021}.
In this paper, we aim to bring photorealism to \frontfacing mobile AR try-on apps, a promising method for empowering consumers with a try-before-purchase experience in e-commerce.  
Our proposed techniques take inspiration from our prior work in lighting estimation~\cite{zhao2020pointar,xihe_mobisys2021,li2021lighting,legendre-facelight} and will address the unique challenges in \frontfacing mobile AR. 
Two baselines we plan to compare against include dual-camera LitAR and EnvMapnet, variants based on their single-camera designs~\cite{zhao2022litar,somanath-envmapnet}.
We expect the first baseline to generate lower-quality environment maps and does not support shadow and the second baseline to output environment maps that do not match the physical environments. 

We believe the key idea of employing multi-camera can open many interesting directions for mobile AR applications. 
This paper focuses on a specific case of \frontfacing and \backfacing dual-camera streaming to improve lighting understanding. 
Our proposed techniques could also benefit other deployment scenarios, such as dual \backfacing cameras and edge-based $360^{\circ}$ cameras. 
As the number of cameras increases, obtaining high-quality environment maps will become easier via coordinated streaming. 
However, questions regarding the energy and quality trade-offs remain unclear and need to be studied further.  

\begin{acks}
We thank the anonymous reviewers for their constructive reviews. 
This work is partly supported by NSF Grants CNS-1815619, NGSDI-2105564, CNS-2236987 and VMWare.
\end{acks}

\balance
\scriptsize{
\bibliographystyle{ACM-Reference-Format}
\bibliography{main}
}

\end{document}